\documentclass{article}

\usepackage{microtype}
\usepackage{graphicx}
\usepackage{subfigure}
\usepackage{amsfonts}
\usepackage{booktabs} 
\usepackage{amssymb}
\usepackage{xcolor}
\pagecolor{white}
\usepackage{hyperref}

\usepackage[accepted]{icml2020}

\icmltitlerunning{Meta-Learning GNN Initializations for Low-Resource Molecular Property Prediction}

\begin{document}

\twocolumn[
\icmltitle{Meta-Learning GNN Initializations for Low-Resource \\
Molecular Property Prediction}




\begin{icmlauthorlist}
\icmlauthor{Cuong Q. Nguyen}{gskai}
\icmlauthor{Constantine Kreatsoulas}{gskdcs}
\icmlauthor{Kim M. Branson}{gskai}
\end{icmlauthorlist}

\icmlaffiliation{gskai}{GlaxoSmithKline, Artificial Intelligence \& Machine Learning}
\icmlaffiliation{gskdcs}{GlaxoSmithKline, Data \& Computational Sciences}
\icmlcorrespondingauthor{Cuong Q. Nguyen}{cuong.q.nguyen@gsk.com}

\icmlkeywords{meta-learning, few-shot learning, graph neural networks, qsar}

\vskip 0.3in
]



\printAffiliationsAndNotice{}  

\begin{abstract}
Building \textit{in silico} models to predict chemical properties and activities is a crucial step in drug discovery. However, limited labeled data often hinders the application of deep learning in this setting. Meanwhile advances in meta-learning have enabled state-of-the-art performances in few-shot learning benchmarks, naturally prompting the question: \emph{Can meta-learning improve deep learning performance in low-resource drug discovery projects?} In this work, we assess the transferability of graph neural networks initializations learned by the Model-Agnostic Meta-Learning (MAML) algorithm -- and its variants FO-MAML and ANIL -- for chemical properties and activities tasks. Using the ChEMBL20 dataset to emulate low-resource settings, our benchmark shows that meta-initializations perform comparably to or outperform multi-task pre-training baselines on 16 out of 20 in-distribution tasks and on all out-of-distribution tasks, providing an average improvement in AUPRC of 11.2\% and 26.9\% respectively. Finally, we observe that meta-initializations consistently result in the best performing models across fine-tuning sets with $k \in \{16, 32, 64, 128, 256\}$ instances.
\end{abstract}

\section{Introduction}
Drug discovery is a multi-parameter optimization process requiring efficient exploration of chemical space for compounds with desired properties. In a typical project, medicinal chemists propose structural changes to compounds in an effort to improve their therapeutic effects without compromising other properties. Validating these changes are costly -- e.g. compounds need to be purchased or synthesized, assays need to be developed and validated -- and thus \textit{in silico} models are often used to prioritize experiments. Following the Merck Molecular Activity Challenge, there has been significant interest in applying deep learning to property prediction. More recently, by directly learning molecular features from chemical graphs, novel architectures in the graph neural networks family have demonstrated improved predictions in quantum chemistry and various property prediction benchmarks \citep{lusci_deep_2013,  duvenaud_convolutional_2015,kearnes_molecular_2016, gilmer_neural_2017, feinberg_potentialnet_2018, yang_analyzing_2019}.

The successes of deep learning, however, hinge on an abundance of data: For instance, ImageNet \citep{deng_imagenet_2009} contains over 14M images and the English Wikipedia database commonly used to pre-train language models has over 2,500M words. On the contrary, labeled scientific data in drug discovery projects often consists of many small, sparse, and heavily biased datasets, consequently limiting the applications of deep learning in this setting. Recent works approach this problem by using pre-training and multitask learning to leverage data from multiple sources \citep{ramsundar_massively_2015,wenzel_predictive_2019, hu_pre-training_2019}.

In parallel, the problem of learning in low-data domain has been tackled vehemently by the few-shot learning community. A prominent solution is the meta-learning paradigm, which aims to learn a learner that is efficient at adapting to new task \citep{thrun_learning_1998,vilalta_perspective_2002,vanschoren_meta-learning_2018}. Matching Networks \citep{vinyals_matching_2016}, a member of this family, have been previously applied to property prediction in one-shot learning settings by \citet{altae-tran_low_2017}. A related approach is the Model-Agnostic Meta-Learning (MAML) algorithm \citep{finn_model-agnostic_2017}, which has been particularly successful at producing state-of-the-arts results on few-shots classification, regression, and reinforcement learning benchmarks, resulting in numerous follow-ups that expand on this elegant framework. 

In this work, we evaluate gated graph neural networks initializations learned by MAML, its first-order approximation (FO-MAML), and the Almost-No-Inner-Loop (ANIL) variant \citep{raghu_rapid_2020} for transfer learning to low-resource molecular properties and activities tasks. Specifically we aim to answer the following questions:
\begin{enumerate}
  \item Does meta-initializations offer improvements over multitask pre-training in this setting?
  \item How little data can meta-initializations learn efficiently from?
\end{enumerate}
Using ChEMBL20 \citep{bento_chembl_2014}, performances of meta-initializations on in- and out-of-distribution tasks are benchmarked with multitask pre-training baselines, showing favorable performances across fine-tuning set sizes of $k \in \{16, 32, 64, 128, 256\}$ instances.

\section{Background}
\textbf{MAML, FO-MAML, and ANIL}\hspace{2mm} MAML's approach to few-shot learning is to directly optimize for a set of initial parameters that is efficient at learning from new data. The algorithm consists of an outer loop that learns an initialization $\theta_{0}$, and an inner loop that adapts $\theta_{0}$ to new tasks. In this setting, a set of tasks $\{T_1, T_2,...,T_K\}$ -- denoted as $\mathcal{T}^{tr}$ -- is available to obtain $\theta_{0}$, from which we would like to learn the set of tasks $\mathcal{T}^{test}$. Following the nomenclature in \citet{finn_model-agnostic_2017}, we call the process of obtaining $\theta_{0}$ \textit{meta-training}, and the process of adapting to $\mathcal{T}^{test}$ \textit{meta-testing}. More formally, we define a task $T_j$ with $K$ instances as $T= \{(x_i, y_i)\,|\, i \in \{1,...,K\}\}$, which is divided into a training set $D^{tr}_{T_j}$ and a test set $D^{test}_{T_j}$, also referred to in the literature as the support and query set, respectively. The inner loop adaptation to $T_j$ for a neural network $f$ parameterized by $\theta$ using gradient descent is expressed as

\[ \theta_N^j = \theta_{N-1}^j - \alpha\nabla_\theta\mathcal{L}_{D^{tr}_{T_j}}(f_{\theta_{n-1}^j})\]
where $\theta_N^j$ denotes the parameters of $f$ after $N$ steps toward task $T_j$, $\alpha$ is the inner loop learning rate, and $\mathcal{L}_{D^{tr}_{T_j}}$ is the loss on the training set of task $T_j$. The loss is calculated using $f$ after $N-1$ updates. The inner loop is repeated for a batch of $B$ tasks sampled from $\mathcal{T}^{tr}$.

For the outer loop, the meta-loss is defined as the sum of task-specific losses after inner loop updates: 
\[ \mathcal{L}_{meta} (\theta_{0}) = \sum_{j=1}^B \mathcal{L}_{D^{test}_{T_j}}(f_{\theta_{N}^j})\]
The task-specific loss $\mathcal{L}_{D^{test}_{T_j}}$ is calculated on the test set of task $T_j$. We then minimize the meta-loss using stochastic gradient descent to optimize the initialization $\theta_{0}$, with updates expressed by
\[ \theta_0 \leftarrow \theta_0 - \eta\nabla_\theta\mathcal{L}_{meta}(\theta_{0})\]
where $\eta$ is the outer loop learning rate. Intuitively, the meta-loss $\mathcal{L}_{meta} (\theta_{0})$ measures how well $\theta_{0}$ adapts to new tasks, and minimizing this loss enables the algorithm to learn good initial parameters. 

Updating $\theta_{0}$ is computationally expensive since it requires the use of second-order derivates to compute $\nabla_\theta\mathcal{L}_{meta} (\theta_{0})$. FO-MAML sidesteps this problem by omitting the second-order terms, effectively ignoring the inner loop gradients. On the other hand, \citet{raghu_rapid_2020} proposes the ANIL algorithm, which reduces the number of second-order gradients required by limiting inner loop adaptation to only the penultimate layer of the network. ANIL and FO-MAML have both demonstrated significant speedup over MAML.

\begin{table}
\caption{Distribution of task types in each split. $A$, $T$, $P$, $B$, and $F$ denote ADME, Toxicity, Physicochemical, Binding, and Functional as found in ChEMBL20. $\mathcal{T}^{tr}$ and $\mathcal{T}^{val}$ only contain $B$ and $F$ tasks, while $\mathcal{T}^{test}$ contains all 5 task types.}
\vskip 0.1in
\label{tab:dataset summary}
\begin{center}
\begin{small}
\begin{sc}
\begin{tabular}{lcccccc} 
\toprule
 & $A$ & $T$ & $P$ & $B$ & $F$ \\
\midrule
$\mathcal{T}^{tr}$ & $0$ & $0$ & $0$ & $126$ & $737$ \\
$\mathcal{T}^{val}$ & $0$ & $0$ & $0$ & $10$ & $10$ \\
$\mathcal{T}^{test}$ & $1$ & $1$ & $1$ & $10$ & $10$ \\
\bottomrule
\end{tabular}
\end{sc}
\end{small}
\end{center}
\vskip -0.1in
\end{table}
\begin{table*}
\caption{Performance on in-distribution tasks measured in AUPRC. The top and bottom halves of the table are tasks with type $B$ and $F$, respectively. Mean and standard deviation are obtained from 25 repeats (see Evaluation in Section \ref{section:exp} for details). The best and second best values are in bold and regular text, respectively. Statistically significant difference from the next best is denoted by ($^*$).}
\label{tab:same distribution}
\begin{center}
\begin{small}
\begin{sc}
\def\arraystretch{1.1}
\resizebox{0.9\textwidth}{!}{%
\begin{tabular}{l  cccccc}
\toprule
ChEMBL ID & k-NN & Finetune-All & Finetune-Top & \textbf{FO-MAML} & \textbf{ANIL} & \textbf{MAML} \\
\midrule
2363236 & $\color{gray}0.316\pm0.007$ & $\color{gray}0.328\pm0.028$ & $\color{gray}0.329\pm0.023$ & $\mathbf{0.337\pm0.019}$ & $\color{gray}0.325\pm0.008$ & $0.332\pm0.013$ \\
1614469 & $\color{gray}0.438\pm0.023$ & $\color{gray}0.470\pm0.034$ & $0.490\pm0.033$ & $\color{gray}0.489\pm0.019$ & $\color{gray}0.446\pm0.044$ & $\mathbf{0.507\pm0.030}$ \\
2363146 & $\color{gray}0.559\pm0.026$ & $0.626\pm0.037$ & $\mathbf{0.653\pm0.029}$ & $\color{gray}0.555\pm0.017$ & $\color{gray}0.506\pm0.034$ & $\color{gray}0.595\pm0.051$ \\
2363366 & $\color{gray}0.511\pm0.050$ & $\color{gray}0.567\pm0.039$ & $\color{gray}0.551\pm0.048$ & $\color{gray}0.546\pm0.037$ & $0.570\pm0.031$ & $\mathbf{0.598\pm0.041}$ \\
2363553 & $\mathbf{0.739\pm0.007}$ & $\color{gray}0.724\pm0.015$ & $0.737\pm0.023$ & $\color{gray}0.694\pm0.011$ & $\color{gray}0.686\pm0.020$ & $\color{gray}0.691\pm0.013$ \\
1963818 & $\color{gray}0.607\pm0.041$ & $0.708\pm0.036$ & $\color{gray}0.595\pm0.142$ & $\color{gray}0.677\pm0.026$ & $\color{gray}0.692\pm0.081$ & $\mathbf{0.745\pm0.048}$ \\
1963945 & $\color{gray}0.805\pm0.031$ & $\mathbf{0.848\pm0.034}$ & $\color{gray}0.835\pm0.036$ & $\color{gray}0.779\pm0.039$ & $\color{gray}0.753\pm0.033$ & $0.836\pm0.023$ \\
1614423 & $\color{gray}0.503\pm0.044$ & $\color{gray}0.628\pm0.058$ & $\color{gray}0.642\pm0.063$ & $0.760\pm0.024$ & $\color{gray}0.730\pm0.077$ & $\mathbf{0.837\pm0.036}^*$ \\
2114825 & $\color{gray}0.679\pm0.027$ & $\color{gray}0.739\pm0.050$ & $\color{gray}0.732\pm0.051$ & $0.837\pm0.042$ & $\color{gray}0.759\pm0.078$ & $\mathbf{0.885\pm0.014}^*$ \\
1964116 & $\color{gray}0.709\pm0.042$ & $\color{gray}0.758\pm0.044$ & $\color{gray}0.769\pm0.048$ & $\color{gray}0.895\pm0.023$ & $0.903\pm0.016$ & $\mathbf{0.912\pm0.013}$ \\
\midrule
2155446 & $\color{gray}0.471\pm0.008$ & $\color{gray}0.473\pm0.017$ & $\color{gray}0.476\pm0.013$ & $0.497\pm0.024$ & $\color{gray}0.478\pm0.020$ & $\mathbf{0.500\pm0.017}$ \\
1909204 & $\color{gray}0.538\pm0.023$ & $\color{gray}0.589\pm0.031$ & $\color{gray}0.577\pm0.039$ & $0.592\pm0.043$ & $\color{gray}0.547\pm0.029$ & $\mathbf{0.601\pm0.027}$ \\
1909213 & $\color{gray}0.694\pm0.009$ & $0.742\pm0.015$ & $\mathbf{0.759\pm0.012}$ & $\color{gray}0.698\pm0.024$ & $\color{gray}0.694\pm0.025$ & $\color{gray}0.729\pm0.013$ \\
3111197 & $\color{gray}0.617\pm0.028$ & $\color{gray}0.663\pm0.066$ & $\color{gray}0.673\pm0.071$ & $\color{gray}0.636\pm0.036$ & $0.737\pm0.035$ & $\mathbf{0.746\pm0.045}$ \\
3215171 & $\color{gray}0.480\pm0.042$ & $\color{gray}0.552\pm0.043$ & $\color{gray}0.551\pm0.045$ & $0.729\pm0.031$ & $\color{gray}0.700\pm0.050$ & $\mathbf{0.764\pm0.019}$ \\
3215034 & $\color{gray}0.474\pm0.072$ & $\color{gray}0.540\pm0.156$ & $\color{gray}0.455\pm0.189$ & $\mathbf{0.819\pm0.048}$ & $\color{gray}0.681\pm0.042$ & $0.805\pm0.046$ \\
1909103 & $\color{gray}0.881\pm0.026$ & $\mathbf{0.936\pm0.013}$ & $0.921\pm0.020$ & $\color{gray}0.877\pm0.046$ & $\color{gray}0.730\pm0.055$ & $\color{gray}0.900\pm0.032$ \\
3215092 & $\color{gray}0.696\pm0.038$ & $\color{gray}0.777\pm0.039$ & $\color{gray}0.791\pm0.042$ & $0.877\pm0.028$ & $\color{gray}0.834\pm0.026$ & $\mathbf{0.907\pm0.017}$ \\
1738253 & $\color{gray}0.710\pm0.048$ & $\color{gray}0.860\pm0.029$ & $\color{gray}0.861\pm0.025$ & $0.885\pm0.033$ & $\color{gray}0.758\pm0.111$ & $\mathbf{0.908\pm0.011}$ \\
1614549 & $\color{gray}0.710\pm0.035$ & $\color{gray}0.850\pm0.041$ & $\color{gray}0.860\pm0.051$ & $0.930\pm0.022$ & $\color{gray}0.860\pm0.034$ & $\mathbf{0.947\pm0.014}$ \\
\midrule\midrule
Avg. Rank & 5.4 & 3.5 & 3.5 & 3.1 & 4.0 & \textbf{1.7} \\
\bottomrule
\end{tabular}}
\end{sc}
\end{small}
\end{center}
\vskip -0.1in
\end{table*}

\begin{table*}[t]
\caption{Performance on out-of-distribution tasks measured in AUPRC. Mean and standard deviations are obtained from 25 repeats (see Evaluation in \ref{section:exp} for details). Notations are the same as Table \ref{tab:same distribution}.}
\label{tab:ood}
\begin{center}
\begin{small}
\begin{sc}
\def\arraystretch{1.1}
\resizebox{0.9\textwidth}{!}{%
\begin{tabular}{l cccccc}
\toprule
ChEMBL ID & k-NN & Finetune-All & Finetune-Top & \textbf{FO-MAML} & \textbf{ANIL} & \textbf{MAML} \\
\midrule
1804798 & $\color{gray}0.338\pm0.020$ & $\color{gray}0.351\pm0.026$ & $\color{gray}0.357\pm0.031$ & $\color{gray}0.360\pm0.017$ & $0.361\pm0.029$ & $\mathbf{0.367\pm0.024}$ \\
2095143 & $\color{gray}0.256\pm0.054$ & $\color{gray}0.147\pm0.046$ & $\color{gray}0.281\pm0.082$ & $0.562\pm0.034$ & $\mathbf{0.564\pm0.037}$ & $\color{gray}0.522\pm0.054$ \\
918058 & $\color{gray}0.407\pm0.138$ & $\color{gray}0.559\pm0.098$ & $0.609\pm0.076$ & $\color{gray}0.506\pm0.096$ & $\color{gray}0.415\pm0.163$ & $\mathbf{0.694\pm0.082}$ \\
\midrule\midrule
Avg. Rank &5.7 & 4.7 & 3.3 & 3.0 & 2.7 & 1.7 \\

\bottomrule
\end{tabular}}
\end{sc}
\end{small}
\end{center}
\end{table*}
\textbf{Graph Neural Networks}\hspace{2mm} The graph neural networks framework enables representation learning on graph structured data by learning node-level representations which are aggregated to form graph-level representations. Throughout our experiments, we use a variant of the Gated Graph Neural Network (GGNN) architecture \citep{li_gated_2017}, a member of the message passing neural network (MPNN) family \citep{gilmer_neural_2017}.
Similar to other MPNNs, the GGNN architecture operates in two phases: a message passing phase and a readout phase. For an undirected graph $\mathcal{G}$ with $V$ nodes where each node has $F$ features, the message passing phase updates the hidden representation of node $v$ at layer $t$ according to
\[ m_v^{t+1} = A_{e_{vv}}h_v^t  + \sum_{w \in N(v)} A_{e_{vw}}h_w^t \]
\[ h_v^{t+1} = \textrm{GRU}(h_v^t ,m_v^{t+1})\]
\vskip 0.05in
where $A_{e_{vw}} \in \mathbb{R}^{F \times F}$ is an edge-specific learnable weight matrix, $N(v)$ denotes neighbors of $v$, GRU is the Gated Recurrent Unit \citep{cho_learning_2014}, and $m_v \in \mathbb{R}^{F}$ is a message used to update the hidden representation of node $v$ denoted by $h_v \in \mathbb{R}^{F}$. Computing the message $m_v$ is often interpreted as aggregating information across central and neighboring nodes. A deviation from \citet{li_gated_2017} comes in our choice to remove weight sharing between GRUs in different layers. Following $T$ updates, the readout phase pools node representations according to
\[ \hat{y} = \textrm{MLP}\bigg(\sum_{v \in G} h_v^T\bigg)\]
to calculate the neural network output $\hat{y}$. Using sum as the readout operation is the second deviation from \citet{li_gated_2017}, and has been shown to have maximal expressive power over mean and max aggregators \citep{xu_how_2018}.

\section{Experimental Settings}
\label{section:exp}
\textbf{ChEMBL20 Dataset}\hspace{2mm} We evaluate the effectiveness of meta-initializations for low-resource tasks using a subset of ChEMBL20. More specifically, the dataset processed by \citet{mayr_large-scale_2018} is filtered for tasks with at least 128 instances. The resulting dataset contains 902 binary classification tasks from 5 distinct task types: ADME ($A$), Toxicity ($T$), Physicochemical ($P$), Binding ($B$), and Functional ($F$).

The tasks are further divided into $\mathcal{T}^{tr}$, $\mathcal{T}^{val}$, and $\mathcal{T}^{test}$. $\mathcal{T}^{val}$ consists of 10 randomly selected $B$ and $F$ tasks. $\mathcal{T}^{test}$ consists of all $A$, $T$, and $P$ tasks in addition to 10 random $B$ and $F$ tasks. The rest of $B$ and $F$ tasks are included in $\mathcal{T}^{tr}$ for meta-training. A summary of task type distribution is shown in Table \ref{tab:dataset summary}. For baselines, $\mathcal{T}^{tr}$ and $\mathcal{T}^{val}$ are combined and split into $D^{tr}_{baseline}$ for training and $D^{val}_{baseline}$ for early stopping. This setup gives the baselines access to more tasks than MAML, FO-MAML, and ANIL during training. 

Each molecule is represented as an undirected graph where nodes and edges are atoms and bonds. We use the OpenEye Toolkit to generate 75 atomic features for each node, similar to those provided by DeepChem \citep{ramsundar_deep_2019}.

\textbf{Baselines}\hspace{2mm} We include the Finetune-All, Finetune-Top, and k-NN baselines as proposed by \citet{triantafillou_meta-dataset:_2019}. All baselines start with training a multi-task GGNN on $D^{tr}_{baseline}$. The k-NN baseline uses the activations from the penultimate layer of pre-trained model to perform classification from 3 nearest neighbors. Finetune-Top reinitializes and trains the penultimate layer while Finetune-All updates all parameters in the model. 

To ensure the baselines are competitive, we perform hyperparameter tuning using the Tree-of-Parzen Estimator implementation of Hyperopt \citep{bergstra_hyperopt_2015} to optimize performance on $D^{val}_{baseline}$. Appendix 1 provides details of the process and the resulting hyperparameters.

\textbf{Meta-Learning}\hspace{2mm} The same GGNN architecture as the baselines is used for all three meta-learning algorithms. Training hyperparameters are hand-tuned for performance on $\mathcal{T}^{val}$ (see Appendix 2 for details). We use the Learn2Learn \citep{arnold_learn2learn_2019} and PyTorch \citep{paszke_pytorch_2019} libraries for our implementation.

\textbf{Evaluation}\hspace{2mm} For each $T_j$ in $\mathcal{T}^{test}$, we fine-tune initializations on $k$ randomly selected instances from $D^{tr}_{T_j}$ using the Adam optimizer with learning rate of 10$^{-4}$ and batch size of $b = min(64, k)$. We use $D^{val}_{T_j}$ for early stopping with patience of 10 epochs and collect performances on $D^{test}_{T_j}$. We use $B$ and $T$ tasks to assess \textit{in-distribution} performance, and $A$, $T$, and $P$ tasks for \textit{out-of-distribution} performance. For each method, the procedure is repeated 25 times with 5 different sets of $k$ instances and 5 random seeds.

\section{Results \& Discussions}
\textbf{Performances on $\mathcal{T}^{test}$}\hspace{2mm} The performance of each method on 23 test tasks when $k = 128$ instances are used for fine-tuning is reported in Table \ref{tab:same distribution} and \ref{tab:ood}. Since random splits have been shown to be overly optimistic in scientific applications \citep{kearnes_modeling_2017, wu_moleculenet_2018}, we emphasize relative ranking over absolute performance throughout our benchmark. We observe that meta-initializations generally exhibit similar or better performances over baselines despite having been trained on fewer tasks. For in-distribution tasks, MAML performs comparably to or outperforms other methods on 16 out of 20 tasks, 2 of which shows significant improvement over the next best method, making it the top performer with an average rank of $1.7$. ANIL and FO-MAML, while benefitting from a shorter training time (Appendix 3), rank $3.1$ and $4.0$ on average, respectively. Similar to observations by \citet{triantafillou_meta-dataset:_2019}, Finetune-All and Finetune-Top baselines prove to be strong competitors, both ranking above ANIL in our benchmark. Given their significantly shorter training time, we suspect both baselines to remain crucial in compute-limited settings. In out-of-distribution settings, meta-initializations outperform baselines on all 3 tasks. Again, MAML is ranked as the best method, followed by ANIL and FO-MAML. Overall, compared to the best baselines, meta-initializations learned by MAML provide an average increase in AUPRC of 11.2\% for in-distribution tasks and 26.9\% out-of-distributions tasks.

\begin{figure}[ht]
\begin{center}
\includegraphics[width=1\linewidth]{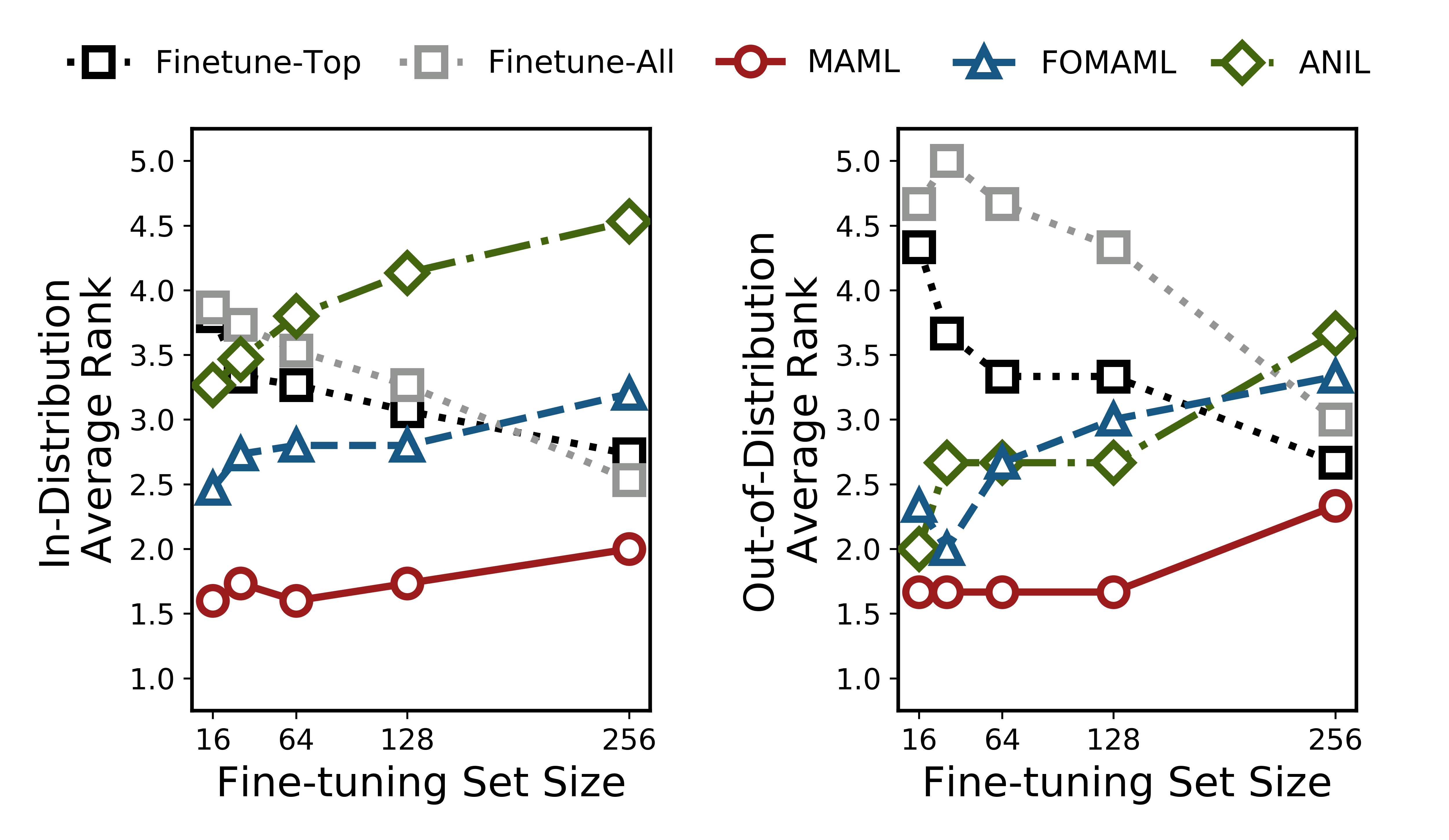}
\vskip -0.1in
\caption{Average ranks of each method performance after fine-tuning with $k \in \{16, 32, 64, 128, 256\}$ instances for in- (left) and out-of-distribution tasks (right). Rankings are based on mean AUPRC measured from five random seeds. MAML is consistently ranked as the best method across all $k$ for in-distribution and out-of-distribution tasks, respectively.}
\label{avg_rank}
\end{center}
\vskip -0.5in
\end{figure}

\textbf{Effect of Fine-tuning Set Size}\hspace{2mm} From $\mathcal{T}^{test}$, we select all tasks with at least 256 instances in $D^{tr}_{T_j}$, resulting in 18 tasks available for evaluation (as opposed to 9 when a threshold of 512 instances is used). The average ranking of each method after fine-tuning on $k \in \{16, 32, 64, 128, 256\}$ instances is reported in Figure \ref{avg_rank} (see Appendix 4 for performance on each task). As the best performing baselines from the previous experiment, Finetune-Top and Finetune-All are selected for comparison. We observe that the baselines benefit greatly from having more data, with Finetune-All rising from fifth to second in in-distribution tasks and Finetune-Top rising from fourth to second in out-of-distribution tasks. Nonetheless, MAML remains the best method, consistently ranked first across fine-tuning set sizes for both sets of tasks.

\section{Conclusion \& Future Directions}
In this work, we explore meta-learning as a tool for learning to predict chemical properties and activities in low-resource settings. Emulating this setting using the ChEMBL20 dataset, we demonstrate that GGNN's initializations learned by MAML perform comparably to or outperform multitask pre-training baselines on 16 out of 20 in-distribution tasks and on all 3 out-of-distribution tasks. Improved performances of meta-initializations are further shown to remain consistent across fine-tuning sets of size $k \in \{16, 32, 64, 128, 256\}$. 

While the ChEMBL20 dataset enables differentiating between in- and out-of-distribution tasks, we recognize that its chemical space is biased towards compounds which have been reviewed and selected for publications. Moreover, our benchmark does not include initializations obtained using self- and un-supervised approaches such as those described in \citet{velickovic_deep_2018}, \citet{hu_pre-training_2019}, and \citet{sun_infograph_2020}. We leave experiments with additional datasets and methods to future work. Overall, we believe our contributions open opportunities in applying deep learning to ongoing drug discovery projects where limited data is available. 

\section*{Acknowledgment}
We would like to thank Jiajie Zhang, Stephen Young, Robert Woodruff, and Darren Green for helpful discussions during the preparation of this manuscript.

\bibliography{metalearning}

\begin{thebibliography}{32}
\providecommand{\natexlab}[1]{#1}
\providecommand{\url}[1]{\texttt{#1}}
\expandafter\ifx\csname urlstyle\endcsname\relax
  \providecommand{\doi}[1]{doi: #1}\else
  \providecommand{\doi}{doi: \begingroup \urlstyle{rm}\Url}\fi

\bibitem[Altae-Tran et~al.(2017)Altae-Tran, Ramsundar, Pappu, and
  Pande]{altae-tran_low_2017}
Altae-Tran, H., Ramsundar, B., Pappu, A.~S., and Pande, V.
\newblock Low {Data} {Drug} {Discovery} with {One}-{Shot} {Learning}.
\newblock \emph{ACS Central Science}, 3\penalty0 (4):\penalty0 283--293, April
  2017.
\newblock ISSN 2374-7943.
\newblock \doi{10.1021/acscentsci.6b00367}.

\bibitem[Arnold et~al.(2019)Arnold, Mahajan, Datta, and
  Ian]{arnold_learn2learn_2019}
Arnold, S. M.~R., Mahajan, P., Datta, D., and Ian, B.
\newblock learn2learn, September 2019.
\newblock URL \url{https://github.com/learnables/learn2learn}.

\bibitem[Bento et~al.(2014)Bento, Gaulton, Hersey, Bellis, Chambers, Davies,
  Krüger, Light, Mak, McGlinchey, Nowotka, Papadatos, Santos, and
  Overington]{bento_chembl_2014}
Bento, A.~P., Gaulton, A., Hersey, A., Bellis, L.~J., Chambers, J., Davies, M.,
  Krüger, F.~A., Light, Y., Mak, L., McGlinchey, S., Nowotka, M., Papadatos,
  G., Santos, R., and Overington, J.~P.
\newblock The {ChEMBL} bioactivity database: an update.
\newblock \emph{Nucleic Acids Research}, 42\penalty0 (D1):\penalty0
  D1083--D1090, January 2014.
\newblock ISSN 0305-1048.
\newblock \doi{10.1093/nar/gkt1031}.

\bibitem[Bergstra et~al.(2015)Bergstra, Komer, Eliasmith, Yamins, and
  Cox]{bergstra_hyperopt_2015}
Bergstra, J., Komer, B., Eliasmith, C., Yamins, D., and Cox, D.~D.
\newblock Hyperopt: a {Python} library for model selection and hyperparameter
  optimization.
\newblock \emph{Computational Science \& Discovery}, 8\penalty0 (1):\penalty0
  014008, July 2015.
\newblock ISSN 1749-4699.
\newblock \doi{10.1088/1749-4699/8/1/014008}.

\bibitem[Cho et~al.(2014)Cho, Merrienboer, Gulcehre, Bougares, Schwenk, and
  Bengio]{cho_learning_2014}
Cho, K., Merrienboer, B.~v., Gulcehre, C., Bougares, F., Schwenk, H., and
  Bengio, Y.
\newblock Learning phrase representations using {RNN} encoder-decoder for
  statistical machine translation.
\newblock \emph{Conference on Empirical Methods in Natural Language Processing
  (EMNLP 2014)}, 2014.

\bibitem[Deng et~al.(2009)Deng, Dong, Socher, Li, Li, and
  Fei-Fei]{deng_imagenet_2009}
Deng, J., Dong, W., Socher, R., Li, L.-J., Li, K., and Fei-Fei, L.
\newblock {ImageNet}: {A} {Large}-{Scale} {Hierarchical} {Image} {Database}.
\newblock In \emph{{CVPR09}}, 2009.

\bibitem[Duvenaud et~al.(2015)Duvenaud, Maclaurin, Aguilera-Iparraguirre,
  Gómez-Bombarelli, Hirzel, Aspuru-Guzik, and
  Adams]{duvenaud_convolutional_2015}
Duvenaud, D., Maclaurin, D., Aguilera-Iparraguirre, J., Gómez-Bombarelli, R.,
  Hirzel, T., Aspuru-Guzik, A., and Adams, R.~P.
\newblock Convolutional {Networks} on {Graphs} for {Learning} {Molecular}
  {Fingerprints}.
\newblock \emph{arXiv:1509.09292 [cs, stat]}, November 2015.
\newblock arXiv: 1509.09292.

\bibitem[Feinberg et~al.(2018)Feinberg, Sur, Wu, Husic, Mai, Li, Sun, Yang,
  Ramsundar, and Pande]{feinberg_potentialnet_2018}
Feinberg, E.~N., Sur, D., Wu, Z., Husic, B.~E., Mai, H., Li, Y., Sun, S., Yang,
  J., Ramsundar, B., and Pande, V.~S.
\newblock {PotentialNet} for {Molecular} {Property} {Prediction}.
\newblock \emph{ACS Central Science}, 4\penalty0 (11):\penalty0 1520--1530,
  November 2018.
\newblock ISSN 2374-7943.
\newblock \doi{10.1021/acscentsci.8b00507}.

\bibitem[Finn et~al.(2017)Finn, Abbeel, and Levine]{finn_model-agnostic_2017}
Finn, C., Abbeel, P., and Levine, S.
\newblock Model-{Agnostic} {Meta}-{Learning} for {Fast} {Adaptation} of {Deep}
  {Networks}.
\newblock In Precup, D. and Teh, Y.~W. (eds.), \emph{Proceedings of the 34th
  {International} {Conference} on {Machine} {Learning}}, volume~70 of
  \emph{Proceedings of {Machine} {Learning} {Research}}, pp.\  1126--1135,
  International Convention Centre, Sydney, Australia, August 2017. PMLR.

\bibitem[Gilmer et~al.(2017)Gilmer, Schoenholz, Riley, Vinyals, and
  Dahl]{gilmer_neural_2017}
Gilmer, J., Schoenholz, S.~S., Riley, P.~F., Vinyals, O., and Dahl, G.~E.
\newblock Neural message passing for {Quantum} chemistry.
\newblock In \emph{Proceedings of the 34th {International} {Conference} on
  {Machine} {Learning} - {Volume} 70}, {ICML}'17, pp.\  1263--1272, Sydney,
  NSW, Australia, August 2017. JMLR.org.

\bibitem[Hu et~al.(2019)Hu, Liu, Gomes, Zitnik, Liang, Pande, and
  Leskovec]{hu_pre-training_2019}
Hu, W., Liu, B., Gomes, J., Zitnik, M., Liang, P., Pande, V., and Leskovec, J.
\newblock Pre-training {Graph} {Neural} {Networks}.
\newblock \emph{arXiv:1905.12265 [cs, stat]}, May 2019.
\newblock arXiv: 1905.12265.

\bibitem[Kearnes et~al.(2016)Kearnes, McCloskey, Berndl, Pande, and
  Riley]{kearnes_molecular_2016}
Kearnes, S., McCloskey, K., Berndl, M., Pande, V., and Riley, P.
\newblock Molecular graph convolutions: moving beyond fingerprints.
\newblock \emph{Journal of Computer-Aided Molecular Design}, 30\penalty0
  (8):\penalty0 595--608, August 2016.
\newblock ISSN 1573-4951.
\newblock \doi{10.1007/s10822-016-9938-8}.

\bibitem[Kearnes et~al.(2017)Kearnes, Goldman, and
  Pande]{kearnes_modeling_2017}
Kearnes, S., Goldman, B., and Pande, V.
\newblock Modeling {Industrial} {ADMET} {Data} with {Multitask} {Networks}.
\newblock \emph{arXiv:1606.08793 [stat]}, January 2017.
\newblock arXiv: 1606.08793.

\bibitem[Kingma \& Ba(2017)Kingma and Ba]{kingma_adam_2017}
Kingma, D.~P. and Ba, J.
\newblock Adam: {A} {Method} for {Stochastic} {Optimization}.
\newblock \emph{arXiv:1412.6980 [cs]}, January 2017.
\newblock arXiv: 1412.6980.

\bibitem[Li et~al.(2017)Li, Tarlow, Brockschmidt, and Zemel]{li_gated_2017}
Li, Y., Tarlow, D., Brockschmidt, M., and Zemel, R.
\newblock Gated {Graph} {Sequence} {Neural} {Networks}.
\newblock \emph{arXiv:1511.05493 [cs, stat]}, September 2017.
\newblock arXiv: 1511.05493.

\bibitem[Lusci et~al.(2013)Lusci, Pollastri, and Baldi]{lusci_deep_2013}
Lusci, A., Pollastri, G., and Baldi, P.
\newblock Deep {Architectures} and {Deep} {Learning} in {Chemoinformatics}:
  {The} {Prediction} of {Aqueous} {Solubility} for {Drug}-{Like} {Molecules}.
\newblock \emph{Journal of Chemical Information and Modeling}, 53\penalty0
  (7):\penalty0 1563--1575, July 2013.
\newblock ISSN 1549-9596.
\newblock \doi{10.1021/ci400187y}.

\bibitem[Mayr et~al.(2018)Mayr, Klambauer, Unterthiner, Steijaert, Wegner,
  Ceulemans, Clevert, and Hochreiter]{mayr_large-scale_2018}
Mayr, A., Klambauer, G., Unterthiner, T., Steijaert, M., Wegner, J.~K.,
  Ceulemans, H., Clevert, D.-A., and Hochreiter, S.
\newblock Large-scale comparison of machine learning methods for drug target
  prediction on {ChEMBL}.
\newblock \emph{Chemical Science}, 9\penalty0 (24):\penalty0 5441--5451, June
  2018.
\newblock ISSN 2041-6539.
\newblock \doi{10.1039/C8SC00148K}.

\bibitem[Paszke et~al.(2019)Paszke, Gross, Massa, Lerer, Bradbury, Chanan,
  Killeen, Lin, Gimelshein, Antiga, Desmaison, Kopf, Yang, DeVito, Raison,
  Tejani, Chilamkurthy, Steiner, Fang, Bai, and Chintala]{paszke_pytorch_2019}
Paszke, A., Gross, S., Massa, F., Lerer, A., Bradbury, J., Chanan, G., Killeen,
  T., Lin, Z., Gimelshein, N., Antiga, L., Desmaison, A., Kopf, A., Yang, E.,
  DeVito, Z., Raison, M., Tejani, A., Chilamkurthy, S., Steiner, B., Fang, L.,
  Bai, J., and Chintala, S.
\newblock {PyTorch}: {An} {Imperative} {Style}, {High}-{Performance} {Deep}
  {Learning} {Library}.
\newblock In Wallach, H., Larochelle, H., Beygelzimer, A., Alché-Buc, F., Fox,
  E., and Garnett, R. (eds.), \emph{Advances in {Neural} {Information}
  {Processing} {Systems} 32}, pp.\  8024--8035. Curran Associates, Inc., 2019.

\bibitem[Raghu et~al.(2020)Raghu, Raghu, Bengio, and Vinyals]{raghu_rapid_2020}
Raghu, A., Raghu, M., Bengio, S., and Vinyals, O.
\newblock Rapid {Learning} or {Feature} {Reuse}? {Towards} {Understanding} the
  {Effectiveness} of {MAML}.
\newblock In \emph{International {Conference} on {Learning} {Representations}},
  2020.

\bibitem[Ramsundar et~al.(2015)Ramsundar, Kearnes, Riley, Webster, Konerding,
  and Pande]{ramsundar_massively_2015}
Ramsundar, B., Kearnes, S., Riley, P., Webster, D., Konerding, D., and Pande,
  V.
\newblock Massively {Multitask} {Networks} for {Drug} {Discovery}.
\newblock \emph{arXiv:1502.02072 [cs, stat]}, February 2015.
\newblock arXiv: 1502.02072.

\bibitem[Ramsundar et~al.(2019)Ramsundar, Eastman, Walters, Pande, Leswing, and
  Wu]{ramsundar_deep_2019}
Ramsundar, B., Eastman, P., Walters, P., Pande, V., Leswing, K., and Wu, Z.
\newblock \emph{Deep {Learning} for the {Life} {Sciences}}.
\newblock O'Reilly Media, 2019.

\bibitem[Sun et~al.(2020)Sun, Hoffmann, Verma, and Tang]{sun_infograph_2020}
Sun, F.-Y., Hoffmann, J., Verma, V., and Tang, J.
\newblock {InfoGraph}: {Unsupervised} and {Semi}-supervised {Graph}-{Level}
  {Representation} {Learning} via {Mutual} {Information} {Maximization}.
\newblock \emph{arXiv:1908.01000 [cs, stat]}, January 2020.
\newblock arXiv: 1908.01000.

\bibitem[Thrun \& Pratt(1998)Thrun and Pratt]{thrun_learning_1998}
Thrun, S. and Pratt, L. (eds.).
\newblock \emph{Learning to {Learn}}.
\newblock Springer US, 1998.
\newblock ISBN 978-0-7923-8047-4.
\newblock \doi{10.1007/978-1-4615-5529-2}.

\bibitem[Triantafillou et~al.(2019)Triantafillou, Zhu, Dumoulin, Lamblin, Evci,
  Xu, Goroshin, Gelada, Swersky, Manzagol, and
  Larochelle]{triantafillou_meta-dataset:_2019}
Triantafillou, E., Zhu, T., Dumoulin, V., Lamblin, P., Evci, U., Xu, K.,
  Goroshin, R., Gelada, C., Swersky, K., Manzagol, P.-A., and Larochelle, H.
\newblock Meta-{Dataset}: {A} {Dataset} of {Datasets} for {Learning} to {Learn}
  from {Few} {Examples}.
\newblock \emph{arXiv:1903.03096 [cs, stat]}, October 2019.
\newblock arXiv: 1903.03096.

\bibitem[Vanschoren(2018)]{vanschoren_meta-learning_2018}
Vanschoren, J.
\newblock Meta-{Learning}: {A} {Survey}.
\newblock \emph{arXiv:1810.03548 [cs, stat]}, October 2018.
\newblock arXiv: 1810.03548.

\bibitem[Veličković et~al.(2018)Veličković, Fedus, Hamilton, Liò, Bengio,
  and Hjelm]{velickovic_deep_2018}
Veličković, P., Fedus, W., Hamilton, W.~L., Liò, P., Bengio, Y., and Hjelm,
  R.~D.
\newblock Deep {Graph} {Infomax}.
\newblock \emph{arXiv:1809.10341 [cs, math, stat]}, December 2018.
\newblock arXiv: 1809.10341.

\bibitem[Vilalta \& Drissi(2002)Vilalta and Drissi]{vilalta_perspective_2002}
Vilalta, R. and Drissi, Y.
\newblock A {Perspective} {View} and {Survey} of {Meta}-{Learning}.
\newblock \emph{Artificial Intelligence Review}, 18\penalty0 (2):\penalty0
  77--95, June 2002.
\newblock ISSN 1573-7462.
\newblock \doi{10.1023/A:1019956318069}.

\bibitem[Vinyals et~al.(2016)Vinyals, Blundell, Lillicrap, kavukcuoglu, and
  Wierstra]{vinyals_matching_2016}
Vinyals, O., Blundell, C., Lillicrap, T., kavukcuoglu, k., and Wierstra, D.
\newblock Matching {Networks} for {One} {Shot} {Learning}.
\newblock In Lee, D.~D., Sugiyama, M., Luxburg, U.~V., Guyon, I., and Garnett,
  R. (eds.), \emph{Advances in {Neural} {Information} {Processing} {Systems}
  29}, pp.\  3630--3638. Curran Associates, Inc., 2016.

\bibitem[Wenzel et~al.(2019)Wenzel, Matter, and
  Schmidt]{wenzel_predictive_2019}
Wenzel, J., Matter, H., and Schmidt, F.
\newblock Predictive {Multitask} {Deep} {Neural} {Network} {Models} for
  {ADME}-{Tox} {Properties}: {Learning} from {Large} {Data} {Sets}.
\newblock \emph{Journal of Chemical Information and Modeling}, 59\penalty0
  (3):\penalty0 1253--1268, March 2019.
\newblock ISSN 1549-9596.
\newblock \doi{10.1021/acs.jcim.8b00785}.

\bibitem[Wu et~al.(2018)Wu, Ramsundar, Feinberg, Gomes, Geniesse, Pappu,
  Leswing, and Pande]{wu_moleculenet_2018}
Wu, Z., Ramsundar, B., Feinberg, E.~N., Gomes, J., Geniesse, C., Pappu, A.~S.,
  Leswing, K., and Pande, V.
\newblock {MoleculeNet}: a benchmark for molecular machine learning.
\newblock \emph{Chemical Science}, 9\penalty0 (2):\penalty0 513--530, January
  2018.
\newblock ISSN 2041-6539.
\newblock \doi{10.1039/C7SC02664A}.

\bibitem[Xu et~al.(2018)Xu, Hu, Leskovec, and Jegelka]{xu_how_2018}
Xu, K., Hu, W., Leskovec, J., and Jegelka, S.
\newblock How {Powerful} are {Graph} {Neural} {Networks}?
\newblock \emph{arXiv:1810.00826 [cs, stat]}, October 2018.
\newblock arXiv: 1810.00826.

\bibitem[Yang et~al.(2019)Yang, Swanson, Jin, Coley, Eiden, Gao, Guzman-Perez,
  Hopper, Kelley, Mathea, Palmer, Settels, Jaakkola, Jensen, and
  Barzilay]{yang_analyzing_2019}
Yang, K., Swanson, K., Jin, W., Coley, C., Eiden, P., Gao, H., Guzman-Perez,
  A., Hopper, T., Kelley, B., Mathea, M., Palmer, A., Settels, V., Jaakkola,
  T., Jensen, K., and Barzilay, R.
\newblock Analyzing {Learned} {Molecular} {Representations} for {Property}
  {Prediction}.
\newblock \emph{Journal of Chemical Information and Modeling}, 59\penalty0
  (8):\penalty0 3370--3388, August 2019.
\newblock ISSN 1549-9596.
\newblock \doi{10.1021/acs.jcim.9b00237}.

\end{thebibliography}
\bibliographystyle{icml2020}

\newpage
\onecolumn
\icmltitle{Appendix: Meta-Learning GNN Initializations for \\
Low-Resource Molecular Property Prediction}
\begin{icmlauthorlist}
\icmlauthor{Cuong Q. Nguyen}{gskai}
\icmlauthor{Constantine Kreatsoulas}{gskdcs}
\icmlauthor{Kim M. Branson}{gskai}

\end{icmlauthorlist}

\vskip 0.3in


\setcounter{section}{0}
\section{Baselines Hyperparameter Tuning}
\label{hyper}
Using Hyperopt, we allow a maximum of 50 evaluations and provide the following search space:
\begin{itemize}
\item[--] Number of GGNN layers: $\{3, 7, 9\}$
\item[--] Fully connected layer dimension: $\{1024, 2048\}$
\item[--] Batch size: $\{128, 256, 512\}$
\item[--] Learning rate: ${10^{\{-4.0,-3.75, -3.5, -3.25\}}}$
\end{itemize}
The resulting architecture has 7 GGNN layers, 1 fully-connected layer with 1024 units, and Dropout applied with a probability of 0.2 at every layer except for the output layer. We use the Adam optimizer with a learning rate of 10$^{-3.75}$, batch size of 512, and patience of 20 epochs for early stopping during pre-training. 
\section{Meta-Learning Hyperparameters}
For MAML and ANIL we use an inner loop learning rate of 0.05, 2 inner gradient steps, and inner batch size of 32, while the outer loop has a learning rate of 0.003 and a batch size of 32. 
FO-MAML uses an outer loop learning rate of 0.0015. 
\section{Training Time} 
\label{Training time}
Training time was measured as the total time required to reach best performance on the $D^{val}_{baseline}$ for baselines and $\mathcal{T}^{val}$ for MAML, FO-MAML, and ANIL on 1 NVIDIA Tesla V100 GPU. We report the recorded times in Table \ref{tab:training time}. The mean and standard deviation are calculated by repeating the training process with five random seeds.
\begin{table}[h]
\caption{Wall clock time to train each method}
\label{tab:training time}
\vskip 0.15in
\begin{center}
\begin{small}
\begin{sc}
\begin{tabular}{l cc } 
\toprule
 & Time (hours) & Speedup\\
\midrule
MAML &$  57.9\pm 0.8 $ & $ 1\times$\\
ANIL & $  48.0\pm 0.6$ & $1.2\times$\\
FO-MAML  &$  27.0\pm 0.9$ & $2.1\times$\\
Multi-task &  $1.4\pm0.1$ & $41.4\times$\\
\bottomrule
\end{tabular}
\end{sc}
\end{small}
\end{center}
\vskip -0.1in
\end{table}
\newpage
\section{Effect of Fine-tuning Set Size on Performances}
\begin{figure}[h]
\vskip 2in
\begin{center}
\includegraphics[width=0.9\linewidth]{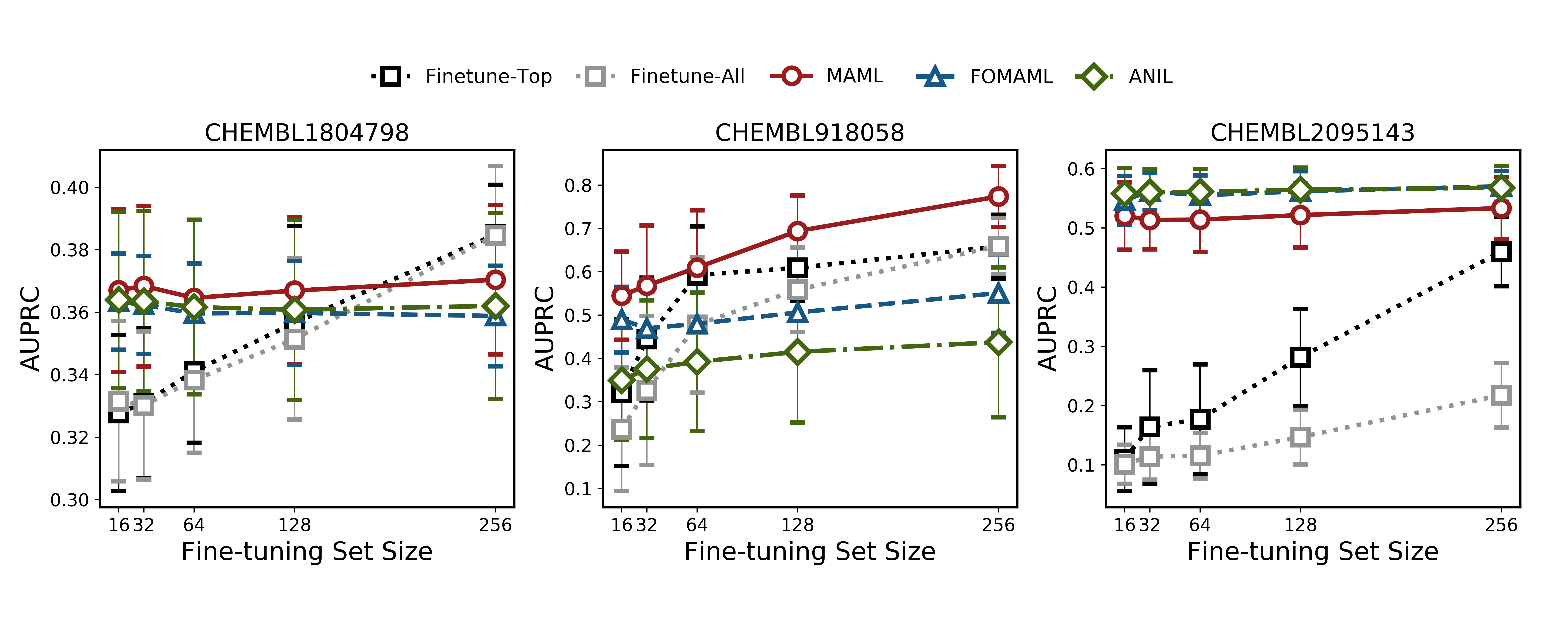}
\caption{Performances on out-of-distribution tasks}
\label{all_out}
\end{center}
\vskip -0.2in
\end{figure}

\begin{figure}
\vskip -0.1in
\begin{center}
\includegraphics[width=0.85\linewidth]{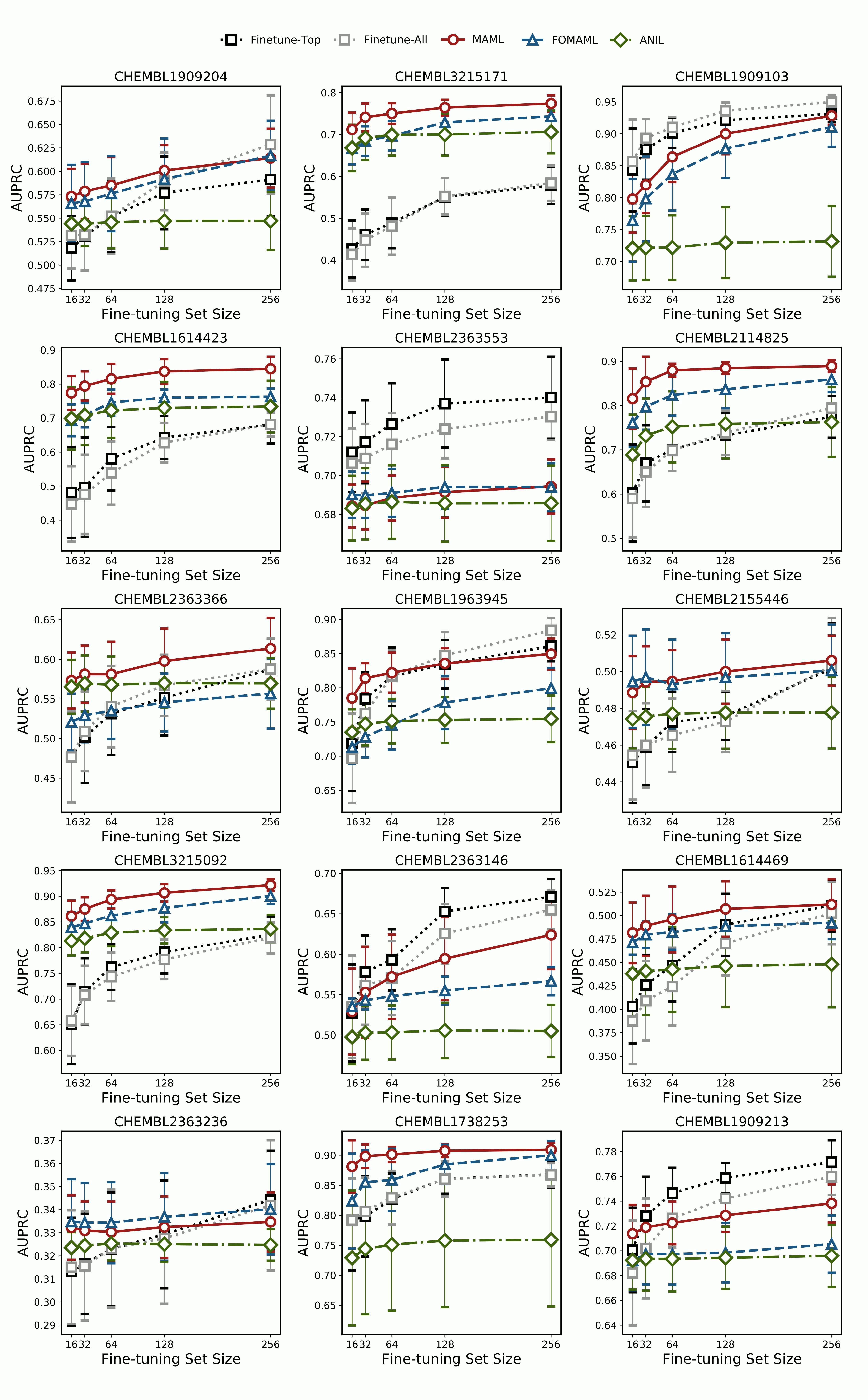}
\vskip -0.3in
\caption{Performances on in-distribution tasks}
\label{all_in_top}

\end{center}
\end{figure}

We report the performances of each method on individual tasks below. Figure \ref{all_in_top} show in-distribution tasks, while Figure \ref{all_out} shows out-of-distribution tasks.
\label{full effect}

\end{document}